\documentclass{article}

\usepackage[preprint]{neurips_2025}

\usepackage[utf8]{inputenc} 
\usepackage[T1]{fontenc}    
\usepackage{hyperref}       
\usepackage{url}            
\usepackage{booktabs}       
\usepackage{amsfonts}       
\usepackage{nicefrac}       
\usepackage{microtype}      
\usepackage{xcolor}         
\usepackage{graphicx}
\usepackage{amsmath} 
\usepackage[ruled,lined,commentsnumbered]{algorithm2e}
\usepackage{xcolor}          
\usepackage{tcolorbox}      
\usepackage{graphicx}
\usepackage{subcaption}
\usepackage{multirow}
\usepackage{booktabs}
\usepackage[labelformat=simple]{caption}
\tcbuselibrary{skins}       
\definecolor{lightpurple}{RGB}{230, 220, 255} 
\newtcolorbox{purplebox}{
  colback=lightpurple,    
  colframe=black,         
  boxrule=1pt,            
  arc=3pt,                
  boxsep=1pt,             
  left=1mm, right=1mm,
  top=1mm, bottom=1mm
}
\title{RLKD: Distilling LLMs' Reasoning via Reinforcement Learning}

\author{Shicheng Xu$^{1,2}$, \; Liang Pang$^{1}$\thanks{Corresponding Author}, \; Yunchang Zhu$^{3}$, \; Jia Gu$^{1,2}$, \; Zihao Wei$^{1,2}$, \; \\ \textbf{Jingcheng Deng$^{1,2}$, \; Feiyang Pan$^{3}$, \;  Huawei Shen$^{1}$, \; Xueqi Cheng$^{1}$ }\\
$^1$State Key Laboratory of AI Safety, Institute of Computing Technology, CAS\\
$^2$University of Chinese Academy of Sciences \; $^3$Huawei Inc. \\
\texttt{\{xushicheng21s,pangliang,shenhuawei,cxq\}@ict.ac.cn}
\\
\texttt{zhuyunchang@huawei.com, pfy824@gmail.com}
}

\begin{document}

\maketitle

\begin{abstract}
Distilling reasoning paths from teacher to student models via supervised fine-tuning (SFT) provides a shortcut for improving the reasoning ability of the smaller Large Language Models (LLMs).
However, the reasoning paths generated by teacher models often reflect only surface-level traces of their underlying authentic reasoning. Insights from cognitive neuroscience suggest that authentic reasoning involves a complex interweaving between meta-reasoning that selects the appropriate sub-problem from multiple candidates, and solving, which addresses the sub-problem. It means that authentic reasoning has implicit multi-branch structure. Supervised fine-tuning collapses this rich structure into a flat sequence of token prediction in teacher's reasoning path, which cannot distill this structure to student. To address this limitation, we propose RLKD, a reinforcement learning (RL)-based distillation framework guided by a novel Generative Structure Reward Model (GSRM). Our GSRM converts the reasoning path into multiple meta-reasoning-solving steps and gives the reward to measure the alignment between the reasoning structures of student and teacher. Our RLKD combines this reward with RL, enables the student LLM to internalize the teacher’s implicit multi-branch structure in authentic reasoning, rather than merely mimicking fixed teacher's output paths. Experiments show that RLKD, even when trained on only 0.1\% of the data under an RL-only regime, surpasses the performance of standard SFT-RL pipelines and further unleashes the potential reasoning ability of the student LLM than SFT-based distillation. Code is available at \url{{https://github.com/xsc1234/RLKD}}.
\end{abstract}

\section{Introduction}

Recently, Large Language Models (LLMs) have demonstrated impressive abilities on complex reasoning tasks~\cite{pan2025survey,jiaqi2025llm,xu2025a} via generating long reasoning paths~\cite{havrilla2024teaching,wei2022chain} such as Deepseek-R1~\cite{guo2025deepseek}. However, high training costs~\cite{guo2025deepseek,xu2025towards} and the strong base model~\cite{chu2025sft,yue2025does} are required in order for LLMs to emerge with this excellent capability, which prevents this reasoning capability from being explored by resource-constrained teams in developing their LLMs. To solve this challenge, supervised fine-tuning (SFT) on the reasoning paths generated from LLMs with powerful reasoning capabilities provides a shortcut and efficient method to make the smaller LLMs generate the long reasoning paths and achieve significant improvement~\cite{guo2025deepseek,zhang2025100,faceopen,wen2025light}. Despite this advance, some studies find SFT-distilled reasoning LLMs are trapped in rigid imitation rather than authentic reasoning. Purely mimicking the teacher’s reasoning paths can leave the student LLM “unthinking”: it replicates the surface form of the reasoning steps yet still makes errors on key underlying steps~\cite{chen2025sft,dai2024beyond}.

\begin{figure*}[t]
    \centering
        \includegraphics[width=1.0\linewidth]{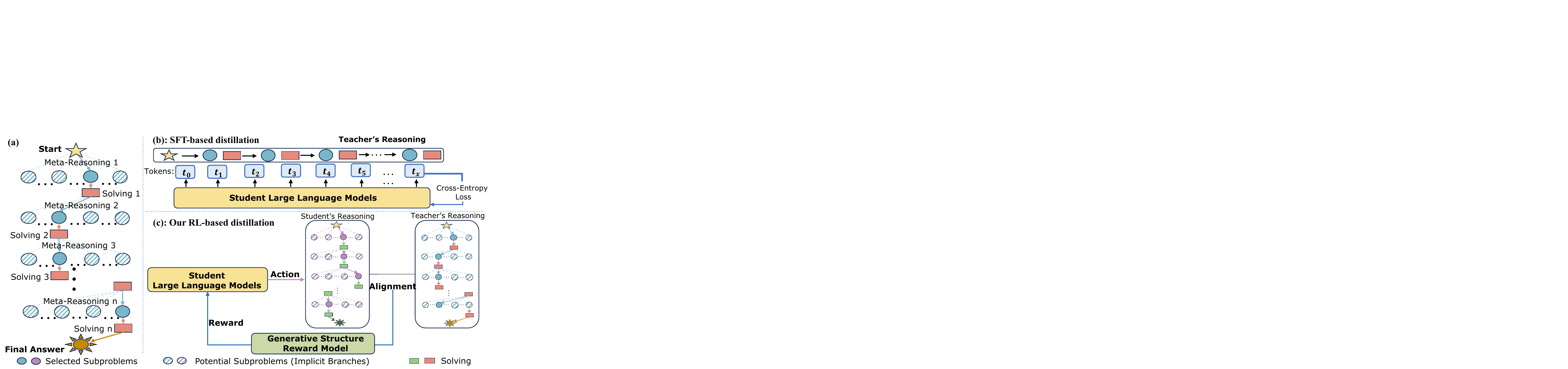}
        \caption{(a) The generated reasoning path has implicit multi-branch structure. (b) Distillation only based on SFT collapses the rich structure into a flat sequence of token prediction to memorize only the teacher's generated path. (c) Our proposed RL-based distillation can teach the student LLM to learn this structure by using a Generative Structure Reward Model to measure the alignment between the reasoning structure of the student and teacher, serving as the reward in RL.}
        \label{motivation}
\end{figure*}

To analyze and solve this phenomenon, we introduce the concepts from human cognitive neuroscience to rethink the definition of authentic reasoning of LLMs, which consists of two parts: one is meta-reasoning and the other is solving~\cite{cox2007metareasoning,russell1991principles}. Specifically, answering a complex problem involves multiple steps and each step consists of a meta-reasoning phase that determines the specific sub-problem from multiple potential sub-problems – followed by a solving phase that executes or answers the specific determined sub-problem. From this perspective, although the generated reasoning content is a definitive path, each step on this path is actually determined by meta-reasoning from multiple candidate states. Therefore, the complex interweaving between meta-reasoning and solving constitutes the authentic reasoning, in which both generated reasoning path and multiple other potential paths form the implicit multi-branch structure (Figure~\ref{motivation} (a)). A critical challenge in SFT-based distillation is that it trains a student LLM to imitate the teacher’s output sequence token-by-token with cross-entropy loss. Therefore, SFT collapses this rich implicit multi-branch structure in teacher into a flat sequence of token prediction to memorize only the teacher's generated path (Figure~\ref{motivation} (b)) while fails to learn how to sample the path from other potential
paths.

To solve this problem, it is important to provide the learning signal in distillation that clearly organizes multi-step meta-reasoning and solving in the reasoning paths, which is the core of implicit multi-branch structure. This is a highly semantic supervision that cannot be accomplished by token-level SFT and requires reinforcement learning (RL)~\cite{ouyang2022training,havrilla2024teaching}, so we propose RLKD, the first reinforcement learning-based knowledge distillation method for LLM's reasoning. In RLKD, we design Generative Structure Reward Model (GSRM), a two-stage reward pattern that combines the semantic understanding of generative reward model~\cite{mahan2024generative} with the interpretability and controllability of the rule-based reward model~\cite{guo2025deepseek}. GSRM can convert the reasoning path into the sequence consisting of multiple meta-reasoning-solving steps, and then score the matching degree between the sequences of the student and the teacher with structured reward mechanism. When combing GSRM with RL, our RLKD can guide the student LLM at a step-level on how to better perform sampling to select the most suitable sub-problem from multiple potential ones , and then solving it. In this way, RLKD can distill the implicit multi-branch structure in the reasoning of teacher LLMs to the student LLMs (Figure~\ref{motivation} (c)).

Experiments on math and graduate-Level Q\&A show that our RLKD: (1) can use only $0.1\%$ training data with RL-only paradigm to outperform SFT-RL pipeline in Qwen2.5-Math, (2) can further unleash the potential reasoning ability of the student LLM than SFT-based distillation, and (3) can outperform existing RL baselines. In conclusion, this paper points out that the authentic reasoning possesses an implicit multi-branch structure, which can not be distilled to the student LLM by SFT and proposes RLKD, which is the first RL-based distillation method for LLM's reasoning. 

  
    
      

\section{Related Work}
\subsection{Structure in Reasoning of LLMs}
Recent studies finds that the reasoning of LLMs has the structure. Chain-of-Thought~\cite{wei2022chain}, Least-to-Most~\cite{zhou2022least} and Self-Ask~\cite{press2022measuring} initially formalize the reasoning path as a chain consisting of multiple nodes, which is a linear structure. Tree-of-Thought~\cite{yao2023tree}, Graph-of-Thought~\cite{besta2024graph} and SearChain~\cite{xu2024search} explicitly builds a non-linear structure for reasoning. SuperCorrect~\cite{yang2025supercorrect} uses high-level plans plus detailed steps as hierarchical thought templates to correct student models. The commonality of existing studies is that they explicitly let LLMs generated the specified structure in reasoning through prompt engineering or fine-tuning. We rethink the reasoning that a generated reasoning path contains multiple potential other paths, and these potential paths together with the generated path form the structure of reasoning. We focus on distilling this implicit multi-branch structure into the student LLM.

\subsection{SFT for Reasoning Distillation}
Supervised fine-tuning (SFT) on chain-of-thought demonstrations has emerged as a straightforward way to distill reasoning capabilities from large models into smaller ones. For example, Deepseek releases a series distilled LLMs based on Deepseek-R1 and significantly improve the reasoning capabilities~\cite{guo2025deepseek}. A similar study shows that with only ~17k curated reasoning traces, a 32B student model can nearly match a closed-source o1-preview on math and coding benchmarks~\cite{li2025llms} and many open-source projects are released~\cite{wen2025light,faceopen,bespoke}. However, recent findings highlight that SFT often teaches format over substance: models learn to imitate the reasoning paths without authentic understanding of their content~\cite{chen2025sft,dai2024beyond}. In fact, a student can produce correct answers by mimicking a long Chain-of-Thought (CoT) pattern even if many intermediate steps are incorrect~\cite{li2025llms}. The key reason behind this is that reasoning has implicit multi-branch structure but SFT distillation collapses this structure into a flat sequence of token prediction. Our RL-based distillation can teach the student LLM to learn this.

\subsection{Reinforcement Learning for LLMs' Reasoning}
Reinforcement learning (RL) has been explored as a means to optimize reasoning strategies in language models, building on foundations like Proximal Policy Optimization (PPO)~\cite{schulman2017proximal}. More recent advances include Group Relative Policy Optimization (GRPO)~\cite{shao2024deepseekmath}, introduced by the DeepSeek team to push mathematical reasoning performance and Decoupled Clip and Dynamic Sampling Policy Optimization (DAPO)~\cite{yu2025dapo} to train LLM with RL at Scale. These methods primarily focus on human feedback, outcome accuracy or heuristic rewards to optimize LLMs in specific downstream tasks. Different from them, our RLKD aims to use RL in knowledge distillation.

\section{Our Method}
This section introduces details of our RLKD, a RL-based knowledge distillation method that can transfer the implicit multi-branch structure for complex reasoning from teacher to student. Firstly, we propose to train a Generative Structure Reward Model (GSRM) to score the alignment degree between the reasoning paths of student and teacher LLMs in terms of their implicit multi-branch structure. Then, we introduce the GSRM into RL-based knowledge distillation.

\subsection{Generative Structure Reward Model}\label{GSRM}
Rewarding the implicit multi-branch structure is premised on accessing the meta-reasoning and solving at each reasoning step. However, the raw reasoning path generated by LLMs such as Deepseek-R1 is unstructured, making it difficult to directly distinguish between various reasoning steps and to decouple the meta-reasoning and solving content of each step. To solve this, we propose Generative Structure Reward Model (GSRM), a two-stage reward pattern that combines the semantic understanding of generative reward model~\cite{mahan2024generative} with the interpretability and controllability of the rule-based reward model~\cite{guo2025deepseek}. In the implementation of GSRM, firstly, we train a LLM to generate a sequence of meta-reasoning and solving pairs for the input reasoning path. Then, we design a structured reward mechanism to score the alignment degree between the teacher and the student in terms of their meta-reasoning and solving content at the corresponding steps.

\begin{figure*}[t]
    \centering
\includegraphics[width=\linewidth]{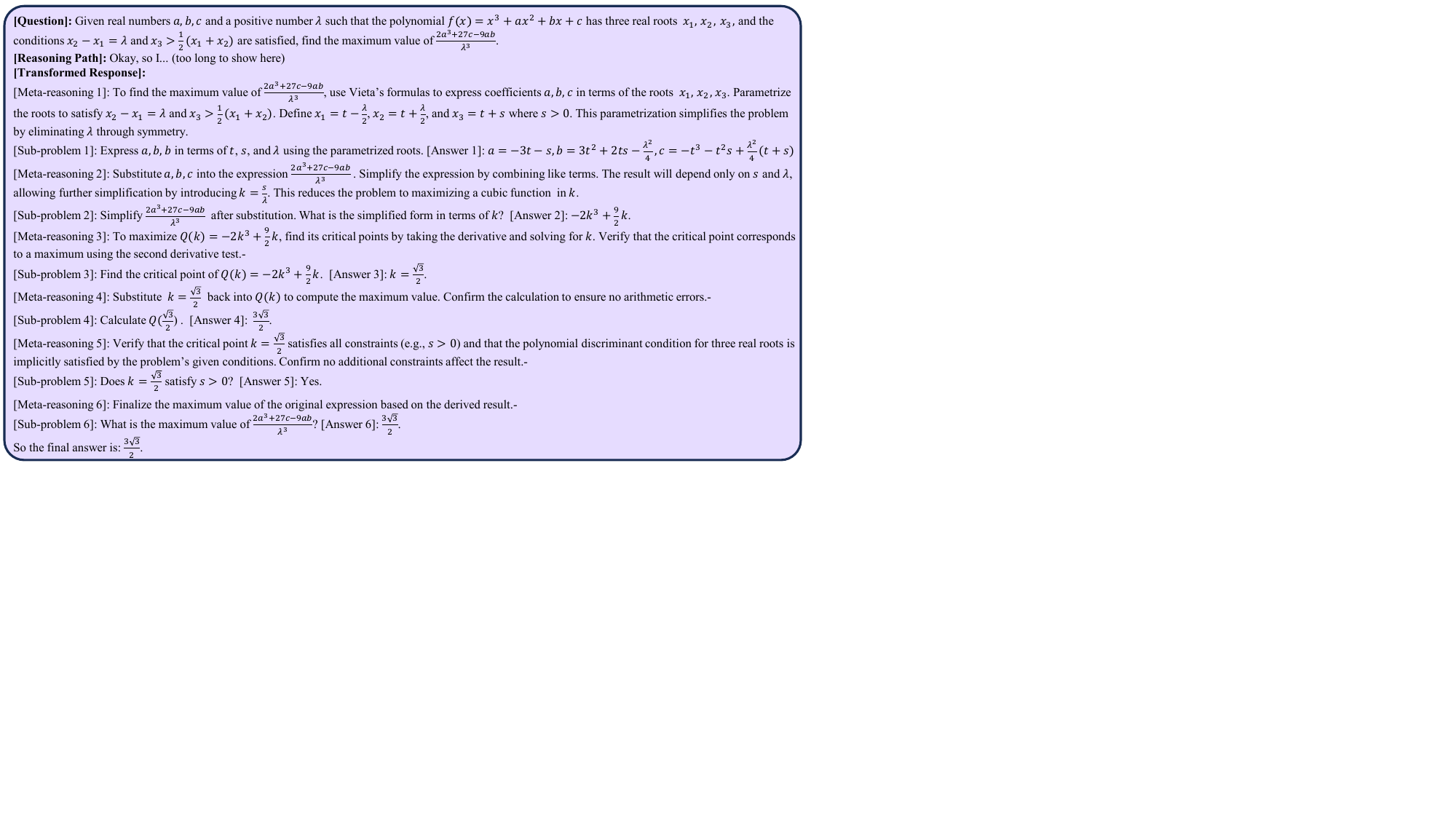}
        \caption{One sequence generation example for math task in in-context learning prompts for GPT-4o.}
        \label{example}
\end{figure*}

\paragraph{Dataset Construction.}
We devise detailed instructions and examples to perform in-context learning (ICL) with the GPT-4o API, enabling it to automatically construct a large-scale supervised fine-tuning dataset. Each data sample of this dataset is a input-out pair: reasoning path $\mathbf{R}$ is the input and the sequence $\mathbf{S}$ consisting of multiple meta-reasoning and solving steps is the output. Specifically, we first structurally define each meta-reasoning and solving steps as ($\mathcal{M}$, $\mathcal{Q}$ \& $\mathcal{A}$), in which $\mathcal{M}$ is the content of meta-reasoning that focuses on determining the sub-problem that the current reasoning step should solve. $\mathcal{Q}$-$\mathcal{A}$ pair is the content of solving, in which $\mathcal{Q}$ is a clear description of the current sub-problem and $\mathcal{A}$ is the solving result to the sub-problem. So sequence $\mathbf{S}$ containing $n$ meta-reasoning and solving steps can be described as:
\begin{align}
    \mathbf{S} = [(\mathcal{M}_1, \mathcal{Q}_1 \& \mathcal{A}_1), (\mathcal{M}_2, \mathcal{Q}_2 \& \mathcal{A}_2), ..., (\mathcal{M}_n, \mathcal{Q}_n \& \mathcal{A}_n)]. \nonumber
\end{align}
The essential for the generation of $\mathbf{S}$ is (1) containing each key reasoning step in the reasoning path and (2) fully decoupling the contents of $\mathcal{M}$, $\mathcal{Q}$, and $\mathcal{A}$ so that they contain and only contain the specified information. Based on these two points, we design specific instruction and examples covering four reasoning tasks (math, science, code and puzzles) to enable GPT-4o to perform effective ICL. One example is shown in Figure~\ref{example} (b).
Besides, we introduce a verification-based feedback strategy to improve data quality. For each $\mathbf{S}$ generated by GPT-4o given input $\mathbf{R}$, we use Deepseek-V3 to determine whether $\mathbf{S}$ meets the requirements of the instruction. If it does not, we give the feedback from Deepseek-V3 to GPT-4o to re-generate $\mathbf{S}$. If a data sample fails to pass the verification after three re-generations, we discard this sample. We automatically execute this data production process on OpenThoughts-114k~\cite{openthoughts}, an open synthetic reasoning dataset with 114k high-quality examples covering math, science, code, and puzzles, and each example has Deepseek-R1 generated reasoning path, which is the $\mathbf{R}$. Finally, we get $93,625$ $\mathbf{R}$-$\mathbf{S}$ pairs for the future supervised fine-tuning. 

\paragraph{Structured Fine-grained Training.} This stage aims to train a generative reward model in our GSRM that can generate the sequence of meta-reasoning and solving pairs for the input reasoning path. Since the target output $\mathbf{S}$ is a highly structured text, we propose a training method called Structured Fine-grained Training to optimize each task (meta-reasoning generation and solving generation) in a fine-grained manner and dynamically adjust the optimization weight according to the difficulty of the task. Specifically, we split the tokens in $\mathbf{S}$ into two sets: one are tokens in meta-reasoning $\mathcal{M}$ and the other are tokens in solving $\mathcal{Q \& A}$. We perform $3$ training epochs. $\mathcal{F}$ is the LLM used in this part (Qwen2.5-7B-Instruct), the first epoch is training for  meta-reasoning, only loss from the tokens in $\mathcal{M}$ is considered:
\begin{align}
    \mathcal{L}_1 = \sum_{\mathbf{S}_{[i]} \in \mathcal{M}} -\log \mathcal{F}(\mathbf{S}_{[i]}|\mathbf{S}_{[1:i-1]}; \theta). \notag
\end{align}
The second epoch is training for solving generation, only loss from $\mathcal{Q \& A}$ is computed as:
\begin{align}
    \mathcal{L}_2 = \sum_{\mathbf{S}_{[i]} \in \mathcal{Q \& A}} -\log \mathcal{F}(\mathbf{S}_{[i]}|\mathbf{S}_{[1:i-1]}; \theta). \notag
\end{align}
The third epoch is mixed training. Total loss is the dynamically weighted average of the losses on the two token sets: 
\begin{align}
    \mathcal{L}_3 & = a\sum_{\mathbf{S}_{[i]} \in \mathcal{M}} -\log \mathcal{F}(\mathbf{S}_{[i]}|\mathbf{S}_{[1:i-1]}; \theta) \\
    &+ b\sum_{\mathbf{S}_{[i]} \in \mathcal{Q \& A}} -\log \mathcal{F}(\mathbf{S}_{[i]}|\mathbf{S}_{[1:i-1]}; \theta). \notag
\end{align}
$a$ and $b$ are initialized to the average of $\mathcal{L}_1$ and $\mathcal{L}_2$ over training steps respectively. During training, we calculate the loss of the tokens in $\mathcal{M}$ and $\mathcal{Q \& A}$ respectively, and update $a$ and $b$ according to the ratio of the losses in two sets every $\alpha$ training steps. We set $\alpha$ as $100$. 

\subsubsection{Structured Reward Mechanism}
After above generation, we denote $\mathbf{S^t}$ and $\mathbf{S^s}$ as the sequences generated from reasoning paths of the teacher and the student LLM respectively:
\begin{align}
        \mathbf{S^t} = [(\mathcal{M}{^t_1}, \mathcal{Q}{^t_1} \& \mathcal{A}{^t_1}), (\mathcal{M}{^t_2}, \mathcal{Q}{^t_2} \& \mathcal{A}{^t_2}), ..., (\mathcal{M}{^t_n}, \mathcal{Q}{^t_n} \& \mathcal{A}{^t_n})], \notag \\
        \mathbf{S^s} = [(\mathcal{M}{^s_1}, \mathcal{Q}{^s_1} \& \mathcal{A}{^s_1}), (\mathcal{M}{^s_2}, \mathcal{Q}{^s_2} \& \mathcal{A}{^s_2}),  ..., (\mathcal{M}{^s_m}, \mathcal{Q}{^s_m} \& \mathcal{A}{^s_m})]. \notag
\end{align}
We propose the Structured Reward Mechanism to map the generated sequence consisting of multiple meta-reasoning and solving to a reward value according to the alignment between $\mathbf{S^t}$ and $\mathbf{S^s}$. Although this computation is carried out based on a linear sequence, when combined with RL, it becomes capable of quantifying the degree of alignment between the teacher and the student in terms of the implicit multi-branch structure, serving as environmental feedback in RL. This is because it can assess the step-level alignment between the sequences and, through the mechanism of reward for RL, finely guide the student LLM to make correct sampling from multiple potential sub-problems at each meta-reasoning step. Compared with SFT-based distillation, it makes the student LLM learn the implicit multi-branch structure in the authentic reasoning of the teacher LLM rather than just focuse on memorizing the reasoning path on the surface generated by the teacher LLM. Compared with existing RL methods, it can guide the student LLM on how to do better sampling at each step and avoid reward hacking~\cite{amodei2016concrete,di2022goal} by step-to-step comparison and early-exist mechanism. Specifically, our structured reward mechanism sequentially compares the corresponding steps of $\mathbf{S^t}$ and $\mathbf{S^s}$ (Algorithm~\ref{alg:reward_calculation_annotated}). We use Qwen-2.5-7B-Instruct to determine whether two texts are matched. For the steps where $\mathcal{M}{^t_i}$  matches $\mathcal{M}{^s_i}$, we assign a temporary reward value of $1$. We further deduct the temporary reward by judging the matching relationship of $\mathcal{Q}_i$ and $\mathcal{A}_i$. The temporary reward value is added to the total reward value after this. When meeting the mismatched $\mathcal{M}_i$, the reward accumulation ends. This reward mechanism follows the sequential dependency of the reasoning path and can give an approximately unique reward value for each step. 
\begin{algorithm}[H]
\caption{Structured Reward Mechanism}
\label{alg:reward_calculation_annotated}
Initialize $r \leftarrow 0$ \tcp*{Total reward accumulator} 
\For{each index $i \in \{1, 2, \ldots, min(n,m)\}$}{
    Initialize temporary reward $v \leftarrow 0$ \tcp*{Step-wise reward score}
    
    \If{Match($\mathcal{M}^{t}_i,\mathcal{M}^{s}_i$)}{
        $v \leftarrow 1$ \tcp*{Base score for matched meta-reasoning}
        
        \If{Not Match($\mathcal{Q}^{t}_i,\mathcal{Q}^{s}_i$)}{
            $v \leftarrow v \times 0.5$ \tcp*{50\% penalty for question mismatch}
        }
        
        \If{Not Match($\mathcal{A}^{t}_i,\mathcal{A}^{s}_i$)}{
            $v \leftarrow v \times 0.5$ \tcp*{Additional 50\% penalty for answer mismatch}
        }
    }
    \Else{Break
        \tcp*{Encountering mismatched meta-reasoning, exit}
    }
    
    $r \leftarrow r + v$ \tcp*{Accumulate step contribution}
}
\Return{$r$} \tcp*{Final reward between sequences}
\end{algorithm}

\subsection{RL-based Knowledge Distillation Training}
We combine our Generative Structure Reward Models (GSRM) with Group-based Relative Policy Optimization (GRPO)~\cite{shao2024deepseekmath} for RL-based knowledge distillation training. In training, we combine the reward obtained from GSRM and the outcome reward of the specific task, such as the accuracy in math, in a weighted manner as the total reward for GRPO. 

\section{Experiments}

\subsection{Experimental Setup.} \label{setup}

\paragraph{Datasets and Evaluation Metrics.}
We use OpenR1-Math as the training dataset for RL, in which Deepseek-R1 generated responses in this dataset are used as the teacher LLM's reasoning paths. We keep the training datasets consistent with baselines including PPO and GRPO. In evaluation, we use the popular and challenging datasets on LLM's reasoning including AIME24~\cite{aime} and MATH-500~\cite{hendrycks2021measuring} for math reasoning and GPQA-Diamond~\cite{rein2024gpqa} for graduate-Level Q\&A. As for the metrics, we follow existing studies on LLM's reasoning~\cite{guo2025deepseek,shao2024deepseekmath,zhang2025100,deng2024everything,xu2024unsupervised} to use pass@k~\cite{chen2021evaluating}. Our validation is divided into two parts: pass@1 and pass@k ($k>1$). As for pass@1, we generate $m$ ($m$ is $64$ for AIME, $8$ for GPQA and $4$ for MATH500) responses for each question and compute pass@1 as $pass@1 = \frac{1}{m}\sum_{i=1}^{m}p_i$, in which $p_i$ is the correctness of the $i$-th response. This can alleviate randomness on small datasets. As for pass@k ($k>1$), it involves the LLM generating $k$ responses for each question, with the data sample being marked as accurate if at least one of the $k$ responses is accurate. Paying attention to the metric where $k>1$ is crucial, as in this setting, the LLM has the opportunity to explore multiple diverse paths to answer the question, it reflects the LLM's ability to sample from multiple implicit paths during reasoning, thereby assessing whether the distilled LLM has learned authentic reasoning or merely memorized the teacher's paths.

\paragraph{Baselines and Comparison Settings.}
We categorize the baselines into three groups according to the experimental settings. The first setting aims to show the effect of our method on improving LLM's reasoning ability. We compare our RL-only training method with the LLM trained in SFT-RL pipeline. We use Qwen2.5-Math-7B-Instruct~\cite{yang2024qwen2}, a powerful reasoning LLM trained on large-scale chain-of-thought datasets with SFT and then GRPO. Both our method and Qwen2.5-Math-7B-Instruct use the same based model: Qwen2.5-Math-7B, which has been pre-trained on math corpus. 
The second setting aims to explore whether our RL-based distillation can further improve the performance of the SFT-distilled LLM and achieve effective optimization of the SFT-RL pipeline in knowledge distillation. We compare our method with RL baselines including PPO and GRPO based on DeepSeek-R1-Distill-Qwen-7B~\cite{guo2025deepseek}, a powerful LLM that is SFT-distilled from Deepseek-R1. The third setting aims to compare which method is better for the student LLM to learn authentic reasoning rather than memorizing the teacher's path. We use embedding\footnote{obtained by gte-Qwen2-7B-instruct} similarity~\cite{deng2025following} to select the 3.2K data samples with the largest difference from the test set in OpenR1-Math-220k training set, and train Qwen2.5-Math-7B on this subset with RL-based distillation and SFT-based distillation respectively.

\paragraph{Implementation Details.}
In training, we build our code based on Open-R1, an open-source project for LLM's reasoning. We use Pytorch 2.5.1 as the training framework and Deepspeed 0.15.4 for acceleration of parallel computing. In RL training, we use one Ascend 910B 64G NPU for online inference and four Ascend 910B 64G NPUs for training under deepspeed-zero3 setting. As for hyperparameters, we set per-device batch size as $2$, gradient accumulation steps as $4$, group size for GRPO as $4$, temperature in online inference as $0.7$. In evaluation, we use Lighteval as the toolkit and follow settings in Deepseek~\cite{guo2025deepseek} to set temperature as $0.6$, max new tokens as $32768$ and top-p as $0.95$. We report the results of multiple runs to reduce the randomness. 

\begin{table*}[t]
  \centering
    \begin{subtable}[t]{\textwidth}
  \renewcommand\arraystretch{1.0}
\setlength\tabcolsep{4pt}
\centering
      \label{tab:sub1}
\scalebox{0.65}{
\begin{tabular}{lcc|cccccccccccc}
\toprule
Model                                     & \multicolumn{2}{c|}{Data Size} & \multicolumn{11}{c}{AIME24}                                                                                                                                 \\ \cmidrule(lr){2-3} \cmidrule(lr){4-9} \cmidrule(l){10-14}
                                          & SFT    & RL     & \multicolumn{6}{c}{pass@1}                                                                 & pass@4 & pass@8 & pass@16 & pass@32 & pass@64 \\
Number of running times for each question    &        &        & 1    & 4    & 8    & 16   & 32   & 64   & 4    & 8    & 16   & 32   & 64   \\ \hline

\multicolumn{14}{c}{\textit{Based on Qwen2.5-Math-7B}}                                                                                                                                                                      \\
Qwen2.5-Math-7B                           & 0      & 0      & 13.3 & 11.7 & 11.7 & 11.3 & 11.0 & 10.3 & 20.0 & 26.7 & 30.0 & 30.0 & 43.3 \\
Qwen2.5-Math-7B-Instruct                  & 2,895K      & 66K      & 16.7 & 15.8 & 15.8 & 15.8 & 14.7 & 14.6 & 33.3 & 43.3 & 43.3 & 50.0 & 50.0 \\
\textbf{Our:} Qwen2.5-Math-7B-RLKD-Zero            & 0      & 3.2K      & 23.3 & 20.0 & 20.4 & 22.1 & 22.5 & 21.0 & 40.0 & 46.7 & 56.7 & 60.0 & 70.0 \\ \hline

\multicolumn{14}{c}{\textit{Based on Deepseek-R1-Distill-Qwen-7B}}                                                                                                                                                          \\
DeepSeek-R1-Distill-Qwen-7B               & 800K      & 0      & 50.0 & 52.7 & 52.5 & 52.9 & 52.3 & 52.4 & 66.7 & 73.3 & 80.0 & 80.0 & 83.3 \\
DeepSeek-R1-Distill-Qwen-7B-PPO           & 800K      & 3.2K      & 46.7 & 52.1 & 53.0 & 52.7 & 52.9 & 52.9 & 66.7 & 73.3 & 73.3 & 80.0 & 83.3 \\
DeepSeek-R1-Distill-Qwen-7B-GRPO          & 800K      & 3.2K      & 50.0 & 52.5 & 53.3 & \textbf{53.3} & 53.3 & 52.3 & 66.7 & 73.3 & 80.0 & 83.3 & 83.3 \\
\textbf{Our:} DeepSeek-R1-Distill-Qwen-7B-RLKD     & 800K      & 3.2K      & \textbf{53.3} & \textbf{56.7} & \textbf{55.4} & \textbf{53.3} & \textbf{52.9} & \textbf{53.6} & \textbf{73.3} & \textbf{80.0} & \textbf{86.7} & \textbf{86.7} & \textbf{86.7} \\ \bottomrule
\end{tabular}
}
      \caption{Performance on AIME24}
    \end{subtable} 
    \begin{subtable}[t]{\textwidth}
  \renewcommand\arraystretch{1.0}
\setlength\tabcolsep{9pt}
\centering
      \label{tab:sub2}
\scalebox{0.65}{
\begin{tabular}{lcc|ccccc|ccc}
\toprule
Model                               & \multicolumn{2}{c|}{Data Size} & \multicolumn{5}{c}{GPQA-Diamond} & \multicolumn{3}{c}{MATH-500} \\ \cmidrule(lr){2-3} \cmidrule(lr){4-6}\cmidrule(lr){7-8} \cmidrule(l){9-10} \cmidrule(l){11-11}
                                    & SFT & RL & \multicolumn{3}{c}{pass@1} & pass@4 & pass@8 & \multicolumn{2}{c}{pass@1} & pass@4 \\
Number of running times for each question            &     &     & 1 & 4 & 8 & 4 & 8 & 1 & 4 & 4 \\ \hline

\multicolumn{11}{c}{\textit{Based on Qwen2.5-Math-7B}} \\
Qwen2.5-Math-7B                     & 0 & 0 & 29.3 & 27.8 & 27.5 & 61.1 & 79.3 & 54.8 & 56.2 & 81.0 \\
Qwen2.5-Math-7B-Instruct            & 2,895K & 66K & 30.3 & 30.2 & 30.1 & 66.7 & 82.3 & 82.4$^*$ & 81.5$^*$ & 89.4$^*$ \\
\textbf{Our:} Qwen2.5-Math-7B-RLKD-Zero      & 0 & 3.2K & 34.9 & 34.2 & 32.7 & 69.2 & 86.4 & 74.4 & 73.9 & 87.8 \\ \hline

\multicolumn{11}{c}{\textit{Based on Deepseek-R1-Distill-Qwen-7B}} \\
DeepSeek-R1-Distill-Qwen-7B         & 800K & 0 & 47.9 & 50.8 & 50.2 & 74.7 & 83.8 & 92.4 & 93.2 & 97.4 \\
DeepSeek-R1-Distill-Qwen-7B-PPO     & 800K & 3.2K & 47.9 & 49.4 & 50.7 & 74.7 & 84.3 & 93.4 & 94.0 & 97.6 \\
DeepSeek-R1-Distill-Qwen-7B-GRPO    & 800K & 3.2K & 50.5 & 50.1 & 49.8 & 75.3 & 84.3 & 93.0 & 93.7 & 97.4 \\
\textbf{Our:} DeepSeek-R1-Distill-Qwen-7B-RLKD               & 800K & 3.2K & \textbf{54.5} & \textbf{53.0} & \textbf{52.8} & \textbf{76.8} & \textbf{86.9} & \textbf{94.2} & \textbf{95.1} & \textbf{98.2} \\ \toprule
\end{tabular}
}
      \caption{Performance on GPQA-Diamond and MATH-500}
    \end{subtable} 
    \small
$^*$means training set of MATH-500 has appeared in the SFT training data of Qwen2.5-Math-7B-Instruct~\cite{yang2024qwen2}.
  \caption{Reasoning abilities on AIME24, MATH-500 and GPQA-Diamond. The results are obtained based on generating multiple responses for each query to mitigate randomness and the best results are in \textbf{bold} font. Qwen2.5-Math-7B-RLKD-Zero is trained by our RLKD without any SFT (RL only).}
    \label{tab:main}
\end{table*}
\subsection{Experimental Results}
\textbf{Main Results. }Results about reasoning abilities of LLMs are shown in Table~\ref{tab:main}. In the training based on Qwen2.5-Math-7B, our RL-only method Qwen2.5-Math-7B-RLKD-Zero outperforms complex SFT + RL pipeline (Qwen2.5-Math-7B-Instruct) and uses much less data (nearly $0.1\%$). In the training based on Deepseek-R1-Distill-Qwen-7B (SFT-distilled LLM), baseline RL methods including PPO and GRPO can hardly bring about significant improvements on this basis while our RLKD is capable of further enhancing performance. The indicates that our RL-based distillation approach enables the SFT-distilled LLM to learn additional information beyond its memorization of the teacher's reasoning paths. The relatively more significant improvements are observed in pass@k ($k>1$). In this setting, LLM has the opportunity to explore multiple diverse paths to answer the question, which means that compared to SFT distillation, our method enables the student LLM to learn how to perform sampling from multiple potential paths by distilling the implicit multi-branch structure from the teacher, thereby increasing the probability of providing the correct answer.

\begin{figure*}[t]
  \centering
  \begin{minipage}[b]{0.33\textwidth}
    \centering
    \includegraphics[width=\textwidth]{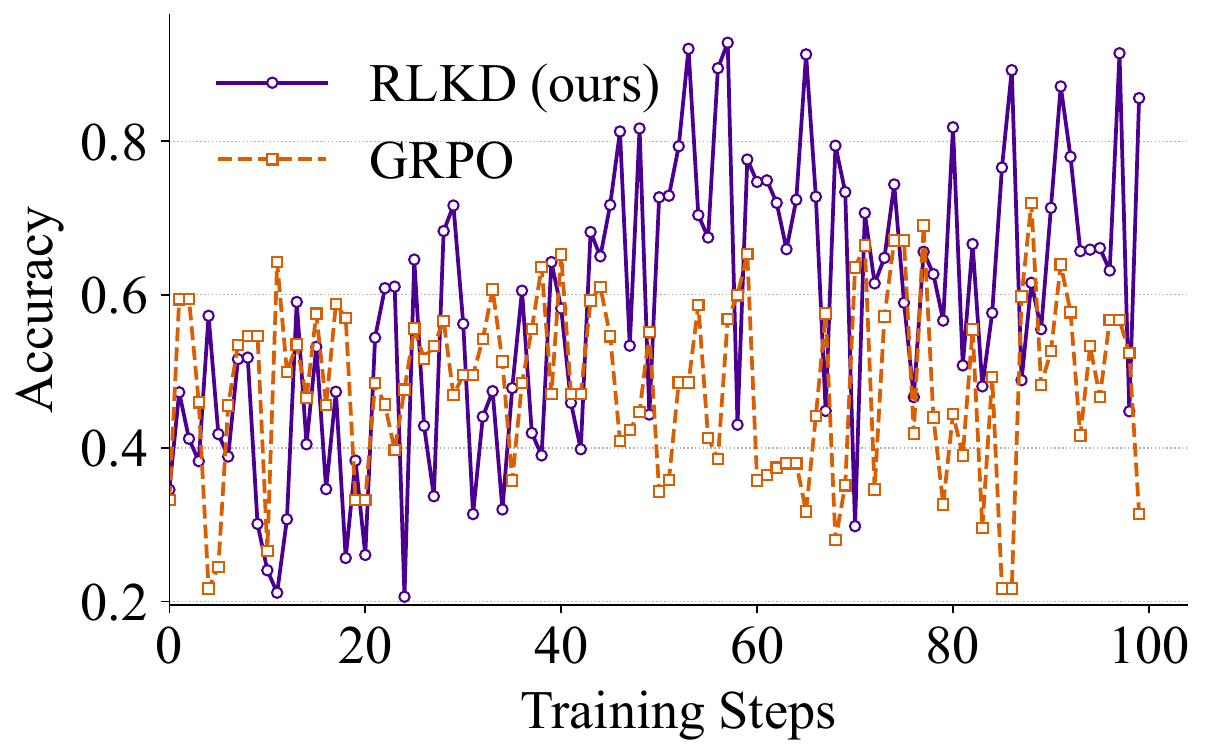}
    \subcaption{Accuracy}
    \label{fig:traina}
  \end{minipage}\hfill%
  \begin{minipage}[b]{0.33\textwidth}
    \centering
    \includegraphics[width=\textwidth]{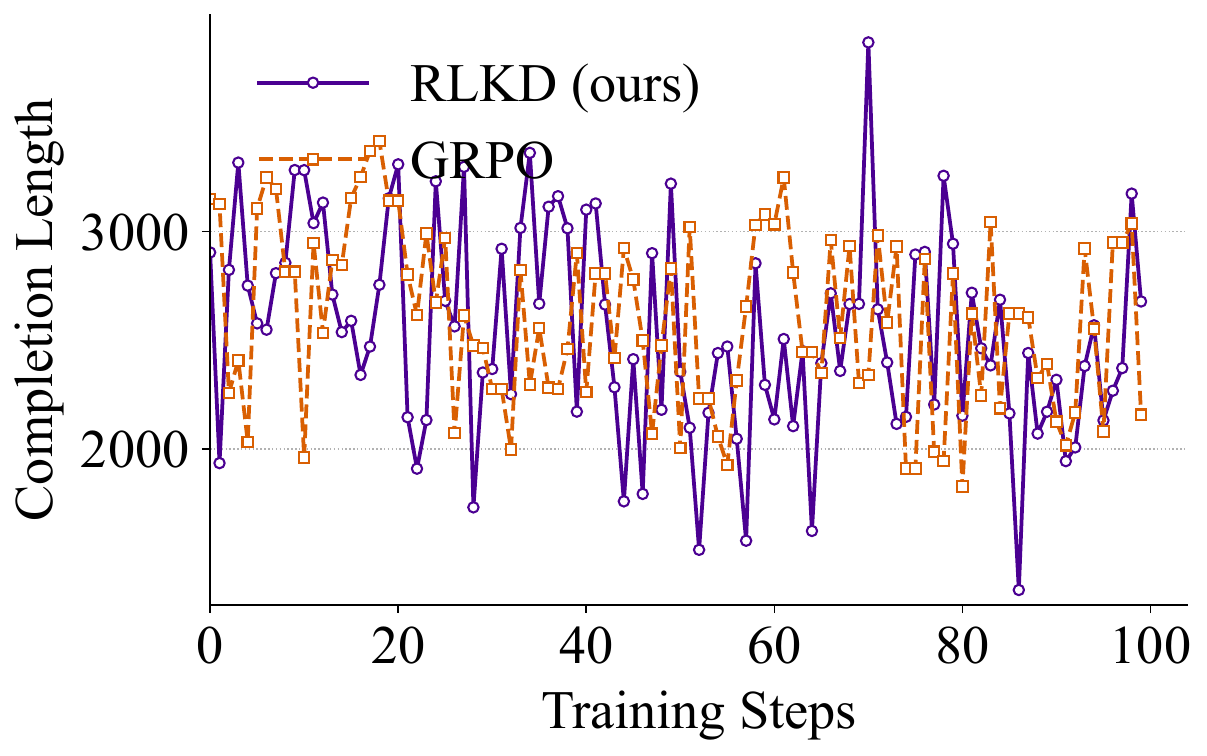}
    \subcaption{Completion Length}
    \label{fig:trainb}
  \end{minipage}\hfill%
  \begin{minipage}[b]{0.33\textwidth}
    \centering
    \includegraphics[width=\textwidth]{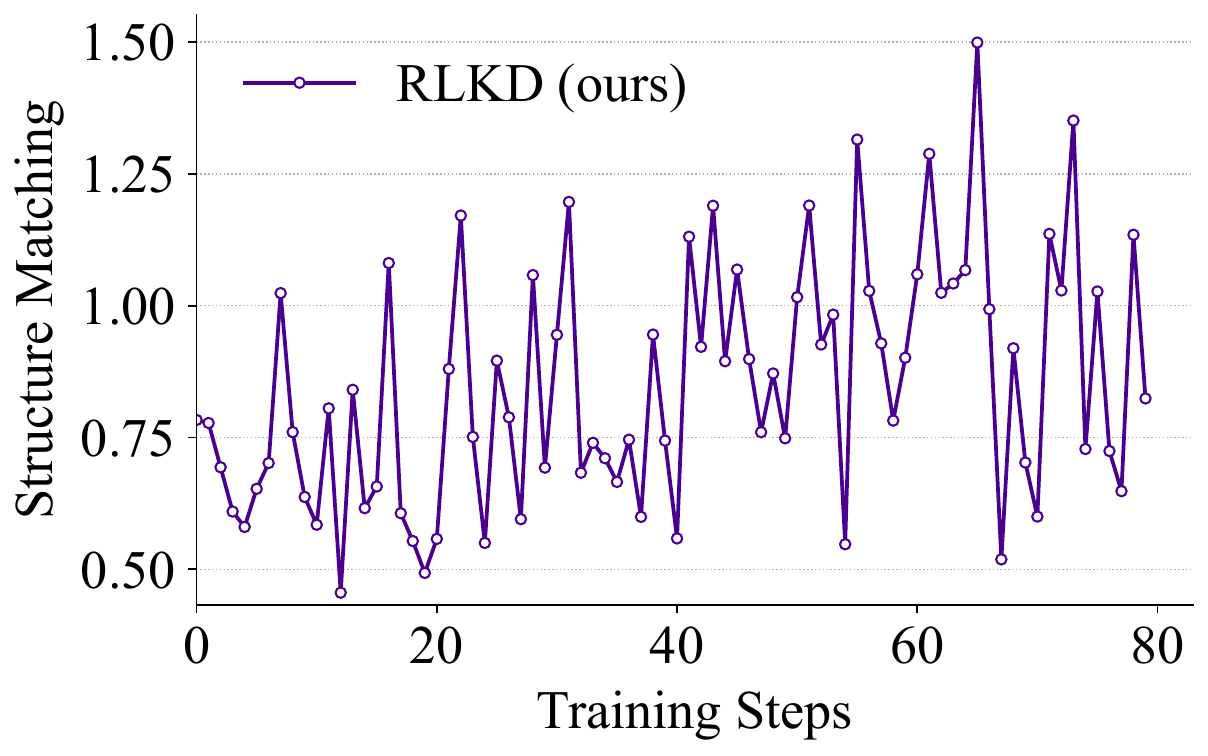}
    \subcaption{Alignment of the Structures}
    \label{fig:trainc}
  \end{minipage}\hfill%

  \caption{The variations of metrics during the RL training process ((c) is not applicable to GRPO).}
  \label{fig:trainall}
\end{figure*}

\textbf{Ablation Study. }Figure~\ref{fig:trainall} shows three metrics in RL training. We compare our RLKD with GRPO because RLKD is actually GRPO with the reward from our Generative Structure Reward Model (GSRM). The results indicate that as the training progresses, GSRM enables RLKD to better optimize the accuracy of the task (Figure~\ref{fig:trainall} (a)), primarily because the student gradually learns the teacher's implicit multi-branch structure in reasoning (Figure~\ref{fig:trainall} (c)). 

\begin{figure*}[t]
  \centering
  \begin{minipage}[b]{0.49\textwidth}
    \centering
    \includegraphics[width=\textwidth]{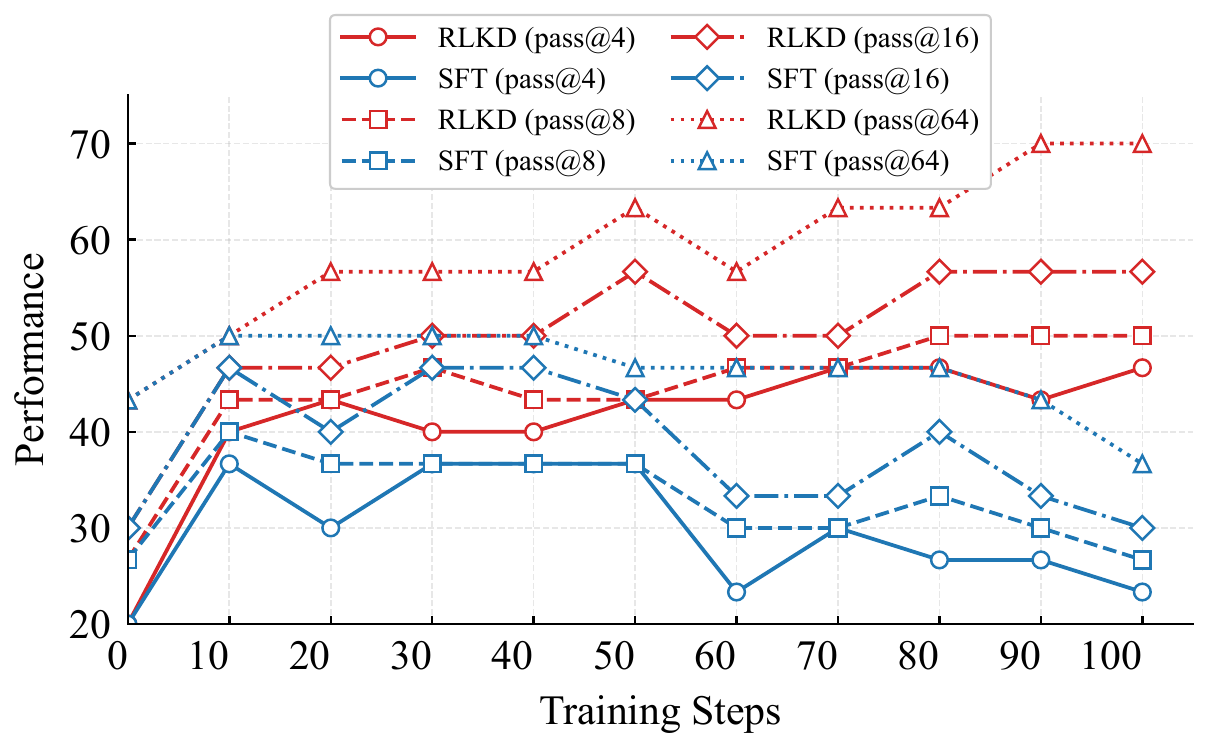}
    \subcaption{AIME24 (Domain Shift)}
    \label{fig:sft-rl-aime}
  \end{minipage}
  \begin{minipage}[b]{0.49\textwidth}
    \centering
    \includegraphics[width=\textwidth]{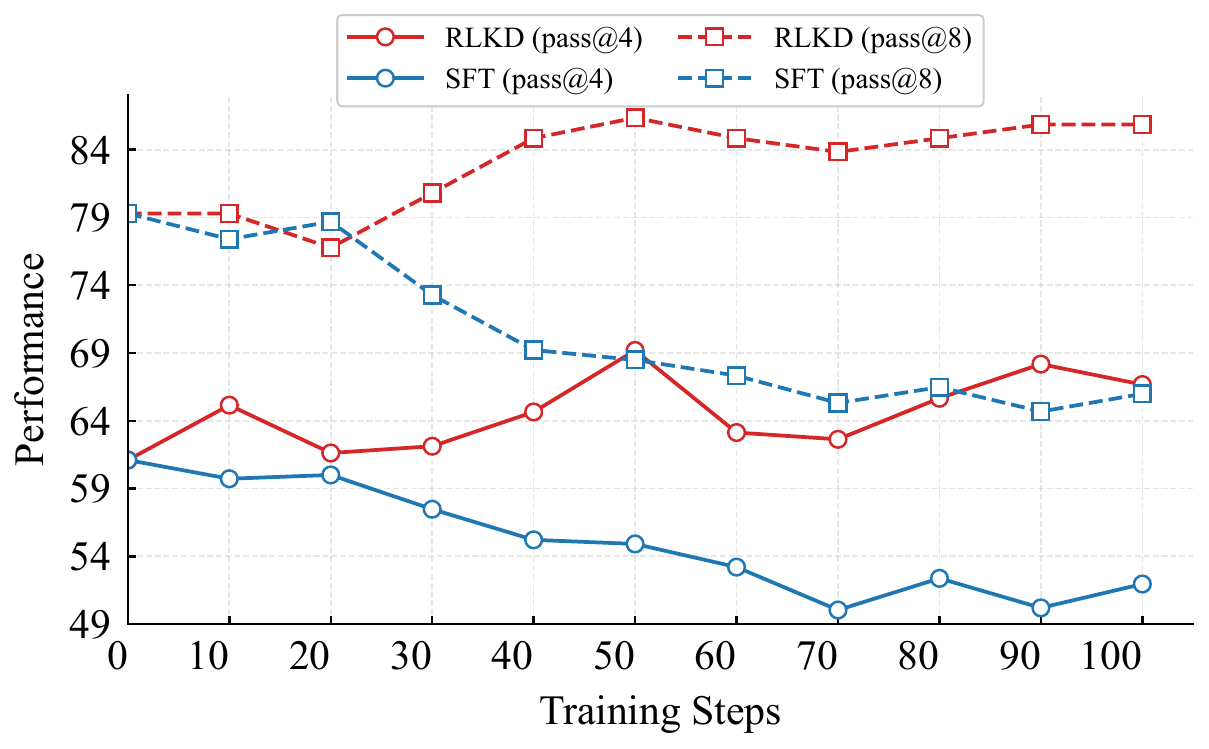}
    \subcaption{GPQA-Diamond (Out-of-Domain)}
    \label{fig:sft-rl-grpo}
  \end{minipage}
  \caption{Comparison between SFT-based distillation and our RL-based distillation (RLKD) in domain shift and out-of-domain setting. SFT and RLKD see the same data (32 samples) at each step.}
  \label{fig:sft-vs-rl}
\end{figure*}

\begin{figure*}[t]
  \centering
  \begin{minipage}[b]{0.19\textwidth}
    \centering
    \includegraphics[width=\textwidth]{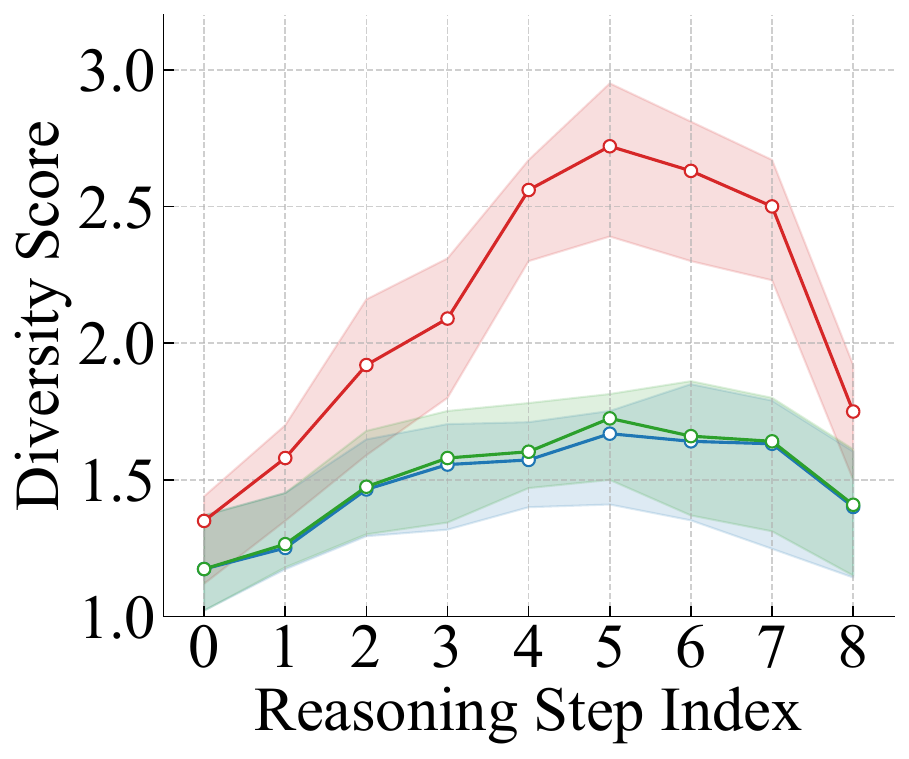}
    \subcaption{Training Step 20}
    \label{fig:a}
  \end{minipage}\hfill%
  \begin{minipage}[b]{0.19\textwidth}
    \centering
    \includegraphics[width=\textwidth]{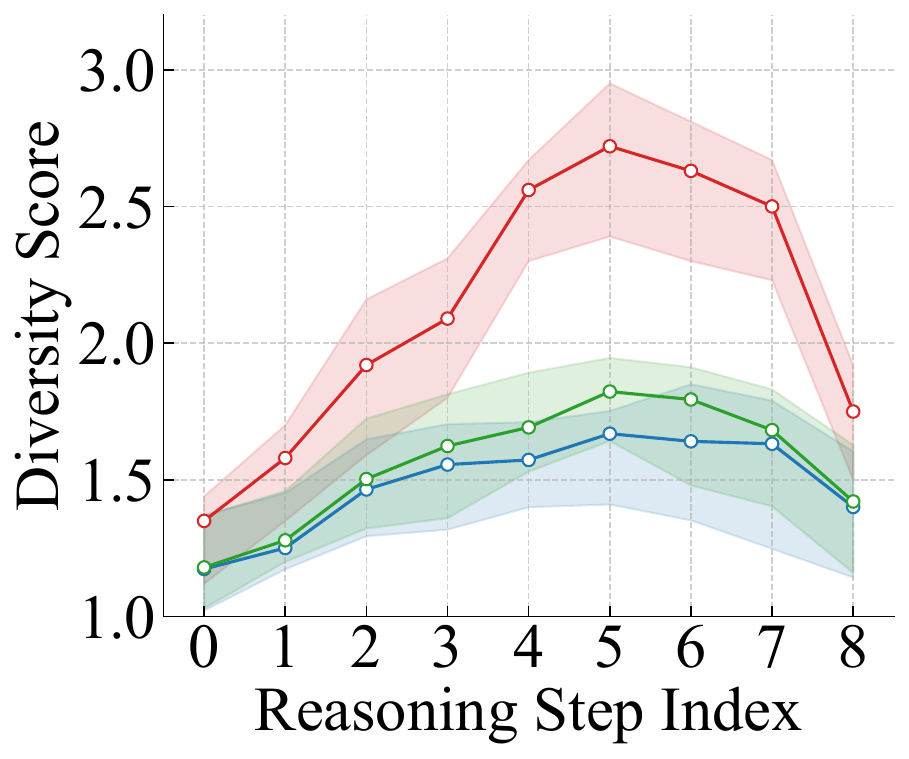}
    \subcaption{Training Step 40}
    \label{fig:b}
  \end{minipage}\hfill%
  \begin{minipage}[b]{0.19\textwidth}
    \centering
    \includegraphics[width=\textwidth]{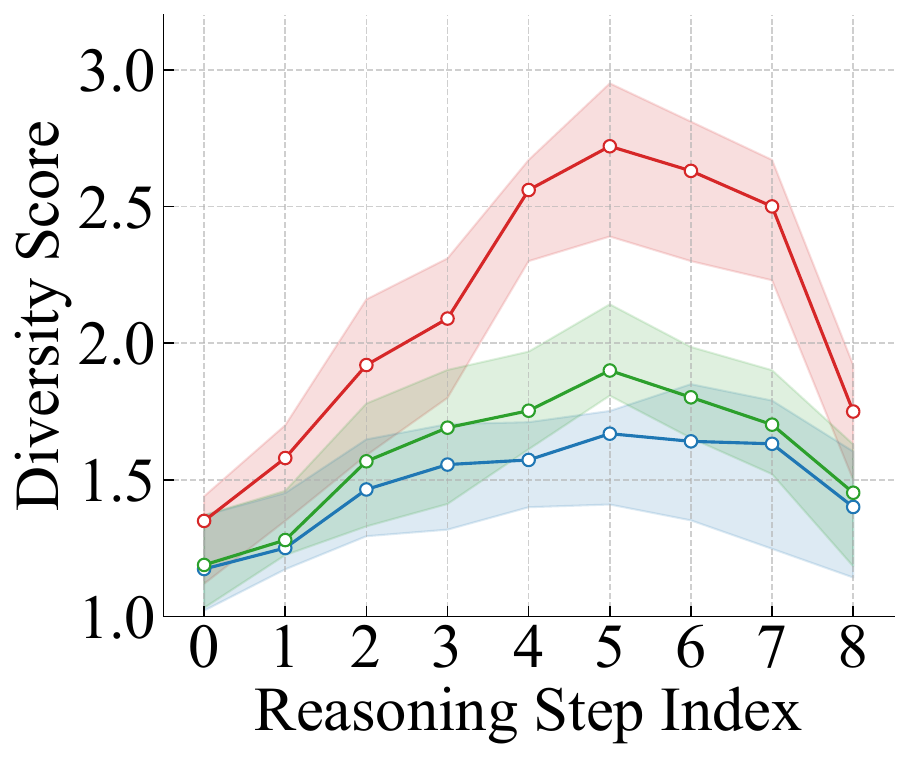}
    \subcaption{Training Step 60}
    \label{fig:c}
  \end{minipage}\hfill%
  \begin{minipage}[b]{0.19\textwidth}
    \centering
    \includegraphics[width=\textwidth]{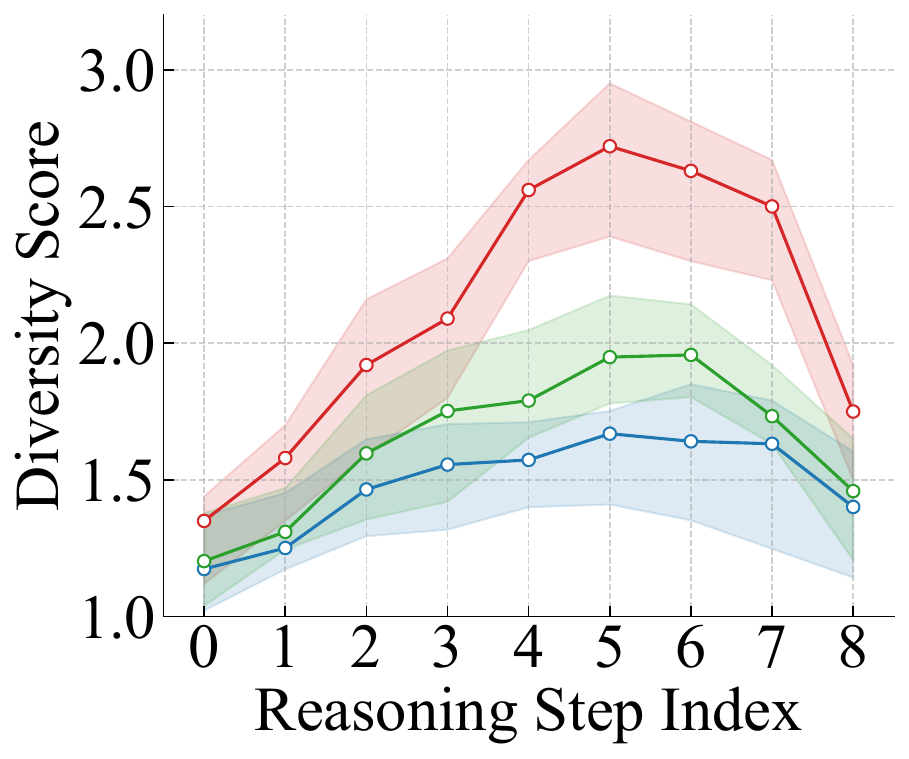}
    \subcaption{Training Step 80}
    \label{fig:d}
  \end{minipage}\hfill%
  \begin{minipage}[b]{0.19\textwidth}
    \centering
    \includegraphics[width=\textwidth]{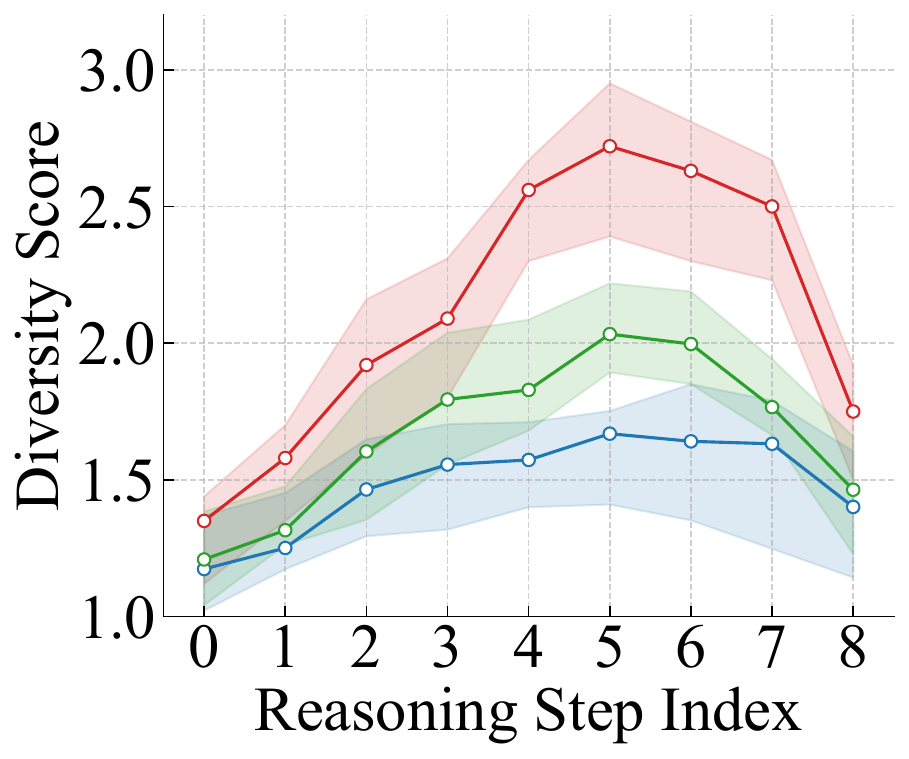}
    \subcaption{Training Step 100}
    \label{fig:e}
  \end{minipage}
  
  \caption{Diversity among different reasoning paths at each meta-reasoning step varying with training. \textcolor{red}{Red} line is Deepseek-R1. \textcolor{blue}{Blue} line is DeepSeek-R1-Distill-Qwen-7B. \textcolor{green}{Green} line is DeepSeek-R1-Distill-Qwen-7B w/ RLKD.}
  \label{fig:all}
\end{figure*}

\subsection{Compare RL Distillation with SFT Distillation}

\paragraph{Performance Trend Varying with Training Steps.}
This section compares the SFT-based distillation with our RL-based distillation (RLKD) by training Qwen2.5-Math-7B on the dataset has domain shift to AIME24 and is out of the domain of GPQA. This setting, where there exists a domain discrepancy between the training and testing sets for distillation, allows us to intuitively discern whether SFT-based distillation is merely mimicking and memorizing the teacher's paths, and whether our RL-based distillation enables the student to learn authentic reasoning. The experimental results are shown in Figure~\ref{fig:sft-vs-rl}. It is noteworthy that, as the training progresses, SFT and RLKD demonstrate completely opposite performance trends: RLKD can consistently enhance performance, even when distilling on dataset that significantly diverges from the testing set, whereas SFT progressively undermines performance. It indicates that SFT distillation is easy to fall into the trap of simply imitating and memorizing the teacher's reasoning paths, rather than learning authentic reasoning that can be stably generalized. Our RLKD teaches students to sample from multiple potential branches, much like the multi-branch structure implicit in real reasoning in teacher LLMs.

\paragraph{Diversity of Reasoning Paths.}
The diversity of reasoning paths can reflect whether the student LLM has truly learned reasoning or is just memorizing the fixed teacher's path in the training set.
Figure~\ref{fig:all} shows that our method brings the diversity-patterns of the student LLM closer to that of the teacher LLM in multi-step reasoning, suggesting that it allows the student LLM to learn authentic reasoning by distilling the implicit multi-branch structure. Specifically, we randomly sample 500 samples from Openthoughts-114K. On this sampled set, we set temperature to $0.8$ and let LLM generate $16$ responses for each question. We use the GSRM in Section~\ref{GSRM} to generate a corresponding sequence containing multiple meta-reasoning-solving steps for each response, and calculate the diversity of the meta-reasoning-solving content within the $16$ responses of each question at step-level. The diversity score $D$ is calculated as:
\begin{align}
\mathbf{u} = \frac{1}{16} \sum_{i=1}^{16} \mathbf{e}_i, \quad
D = \frac{1}{\frac{1}{16} \sum_{i=1}^{16} \frac{\mathbf{e}_i \cdot \mathbf{u}}{|\mathbf{e}_i| |\mathbf{u}|}}, \notag
\end{align}
\noindent in which $\mathbf{e}_i$ is the text embedding encoded by \textit{gte-Qwen2-7B-instruct} for each meta-reasoning-solving content. As shown in Figure~\ref{fig:all}, the teacher LLM (Deepseek-R1) has a significantly different diversity-pattern from the SFT-distilled student LLM (DeepSeek-R1-Distill-Qwen-7B). Reasoning in the teacher has the higher diversity while the student is stuck in the relatively fixed paths. As the training of our method progresses (from step 20 to 100), the diversity of the student’s reasoning paths begins to increase and gradually approaches that of the teacher, which indicates that our method allows the student to learn the teacher’s authentic reasoning paradigm.

\section{Conclusion and Discussion} \label{con_dis}
This work addresses a critical flaw in knowledge distillation in LLM's reasoning: the failure of SFT to transfer the implicit multi-branch structure underlying authentic reasoning. Drawing from cognitive neuroscience principles, we show that authentic reasoning involves dynamic meta-reasoning (sub-problem selection) and solving steps—a implicit multi-branch structure flattened by token-level SFT training. Our RLKD, the first RL-based distillation framework for reasoning, overcomes this paired with a Generative Structure Reward Model (GSRM), which decomposes reasoning paths into meta-reasoning-solving pairs and scores the structural alignment between teacher and student. Experiments across math and graduate-level QA tasks demonstrate RLKD’s superiority than SFT-based distillation, SFT-RL pipeline and RL baselines including PPO and GRPO, proving its ability to distill how teachers navigate latent reasoning branches rather than mimicking surface tokens. Further analysis confirms RLKD-trained students mirror teachers’ multi-branch exploration patterns, closing the imitation-authentic reasoning gap. 


\section*{Acknowledgments}
This work was supported by the Key Research and Development Program of Xinjiang Uyghur Autonomous Region Grant No. 2024B03026, the Strategic Priority Research Program of the CAS under Grants
No.XDB0680302, the Beijing Nova Program
under Grants No. 20250484765, the National Natural Science Foundation of China (NSFC) under Grants No.
62276248, and the Youth Innovation Promotion Association CAS under Grants
No. 2023111.

\bibliographystyle{unsrtnat}
\bibliography{neurips_2025}


\appendix

\section{Ablation study of Structured Fine-grained Training}
\begin{figure}[h]
  \centering
  \begin{minipage}[b]{0.49\textwidth}
    \centering
    \includegraphics[width=\textwidth]{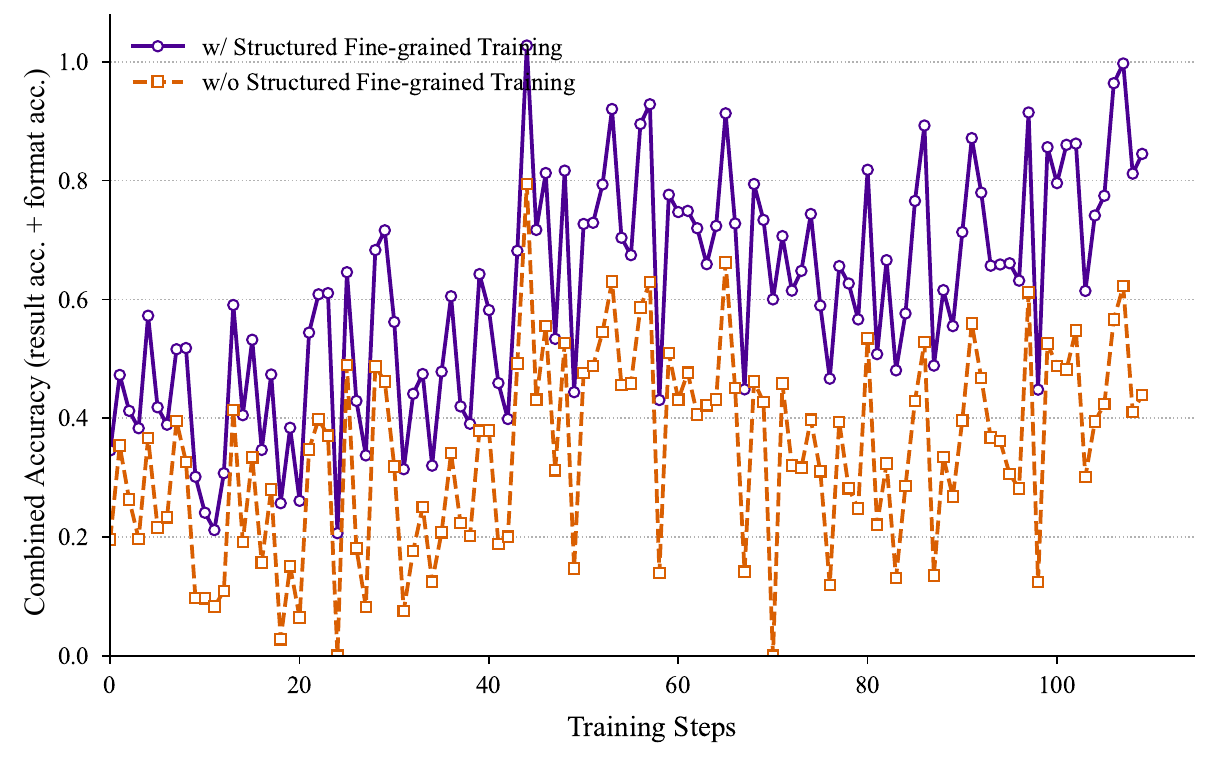}
    \subcaption{Combined Accuracy (result acc. + format acc.)}
    \label{fig:ablationa}
  \end{minipage}\hfill%
  \begin{minipage}[b]{0.49\textwidth}
    \centering
    \includegraphics[width=\textwidth]{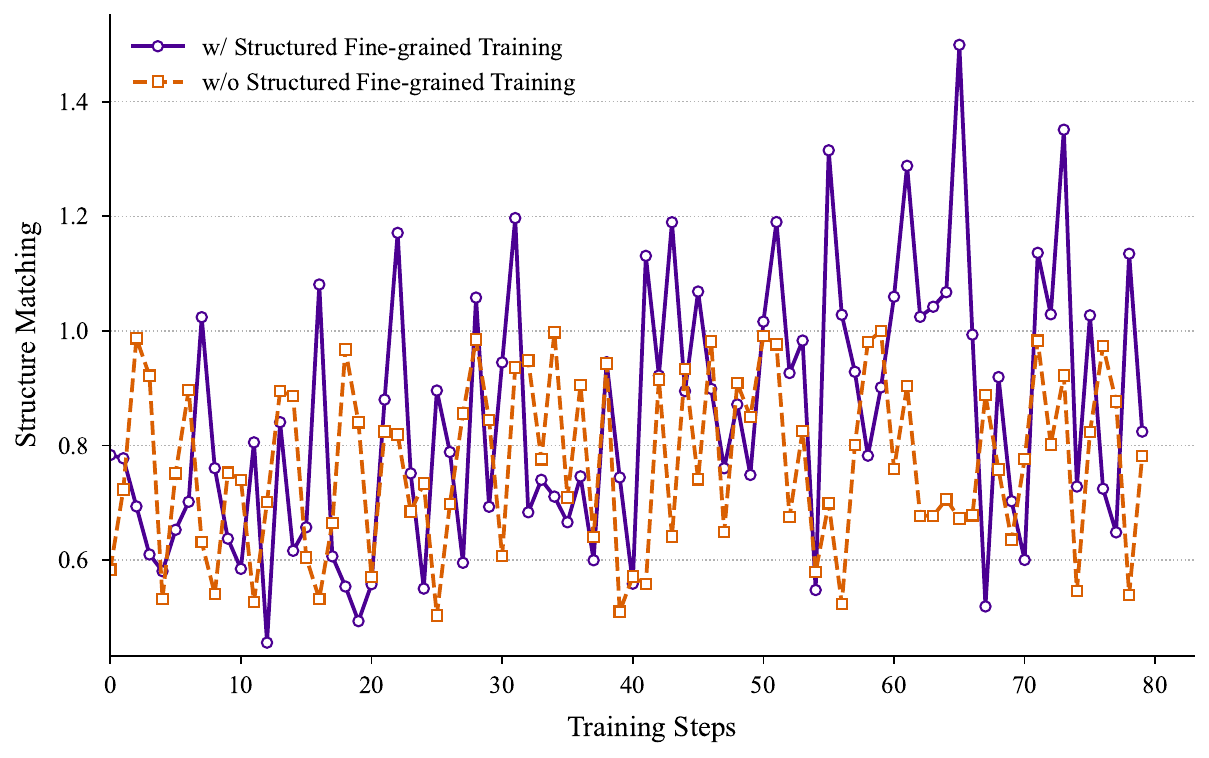}
    \subcaption{Structure Matching}
    \label{fig:ablationb}
  \end{minipage}\hfill%

  \caption{Ablation study about our Structured Fine-grained Training method.}
  \label{fig:ablationall}
\end{figure}

Figure~\ref{fig:ablationall} shows the ablation study of our Structured Fine-grained Training method by showing the accuracy and structure matching varying with training steps. It indicates that generative structure reward model trained by our structured fine-grained training method can effectively optimize the task accuracy (Figure~\ref{fig:ablationa}) and the matching degree of reasoning structure (Figure~\ref{fig:ablationb}) between the teacher and the student, which grows progressively as the training steps.

\section{Ablation study for Reward Weights}

In this section, we conduct ablation study about the reward weights in RL training. The essential rewards in our method include accuracy reward ($R_{acc}$) for the specific tasks such as rule-based accuracy reward in math, and our structure matching reward ($R_{gsrm}$) obtained from generative structure reward model, so we primarily discuss the weight relationship between these two rewards. Specifically, we adjust the weight between $R_{gsrm}$ and $R_{acc}$ ($\frac{R_{gsrm}}{R_{acc}}$) from $0.2$ to $2$ in increments of $0.2$, and observe the variations in pass@1 (n=16) and pass@16 (n=16) on AIME24. This experimental results are shown in Figure~\ref{fig:ablation-reward-weights}. We can see that when the weights of the two are equal, i.e., $\frac{R_{gsrm}}{R_{acc}} = 1$, the LLM achieves a harmonized optimal effect in both the correctness of reasoning (pass@1)and the capability of sampling (pass@16). So in our experiments, we set the reward weights as: $ R_{acc}:R_{gsrm}:R_{format}:R_{tag} = 3:3:2:2$, in which $R_{format}$ is format reward and $R_{tag}$ is tag count reward that assesses whether the thought content adheres to the specified format~\cite{guo2025deepseek}.

\begin{figure}[t]
  \centering
  \begin{minipage}[b]{0.49\textwidth}
    \centering
    \includegraphics[width=\textwidth]{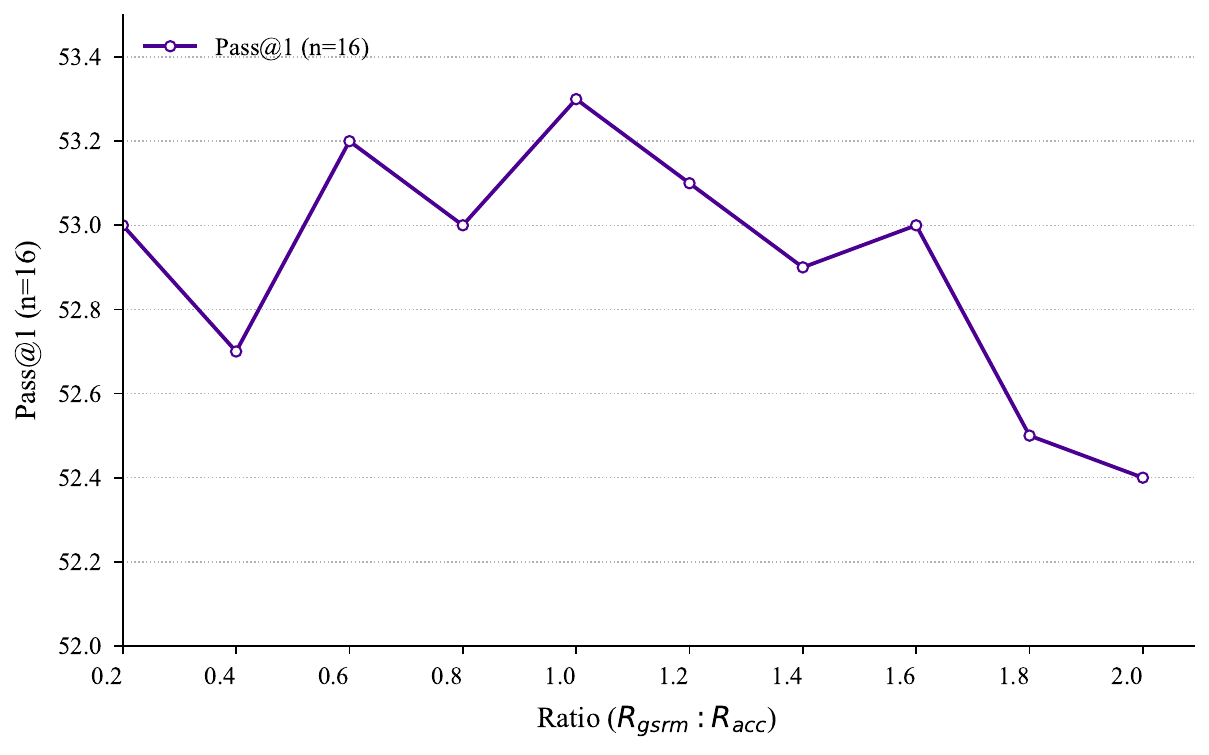}
    \subcaption{pass@1 (n=16) on AIME24}
    \label{fig:ablation-reward-weights}
  \end{minipage}\hfill%
  \begin{minipage}[b]{0.49\textwidth}
    \centering
    \includegraphics[width=\textwidth]{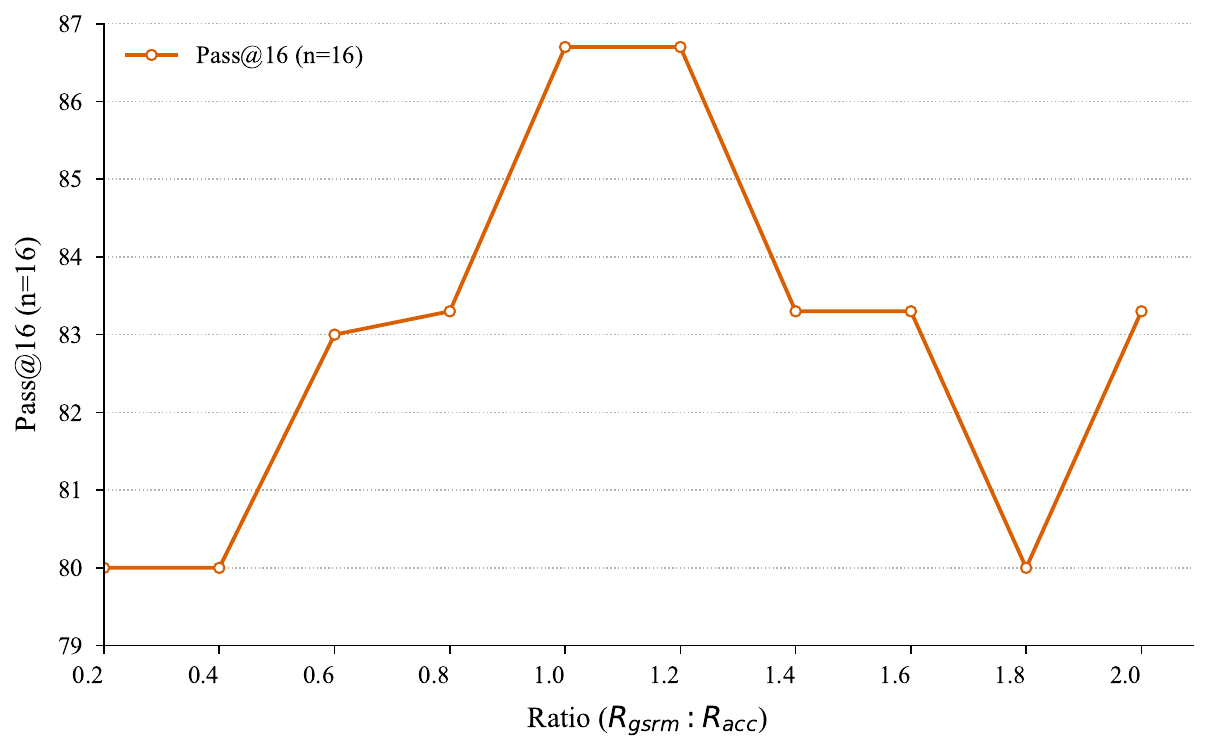}
    \subcaption{pass@16 (n=16) on AIME24}
    \label{fig:ablation-reward-weights}
  \end{minipage}\hfill%

  \caption{Ablation study about reward weights. Performance of pass@1 (n=16) and pass@16 (n=16) on AIME24 varying with ($\frac{R_{gsrm}}{R_{acc}}$).}
  \label{fig:ablation-reward-weights}
\end{figure}

\section{Prompt in Data Construction}

\begin{figure}[h]
    \centering
        \includegraphics[width=1.0\linewidth]{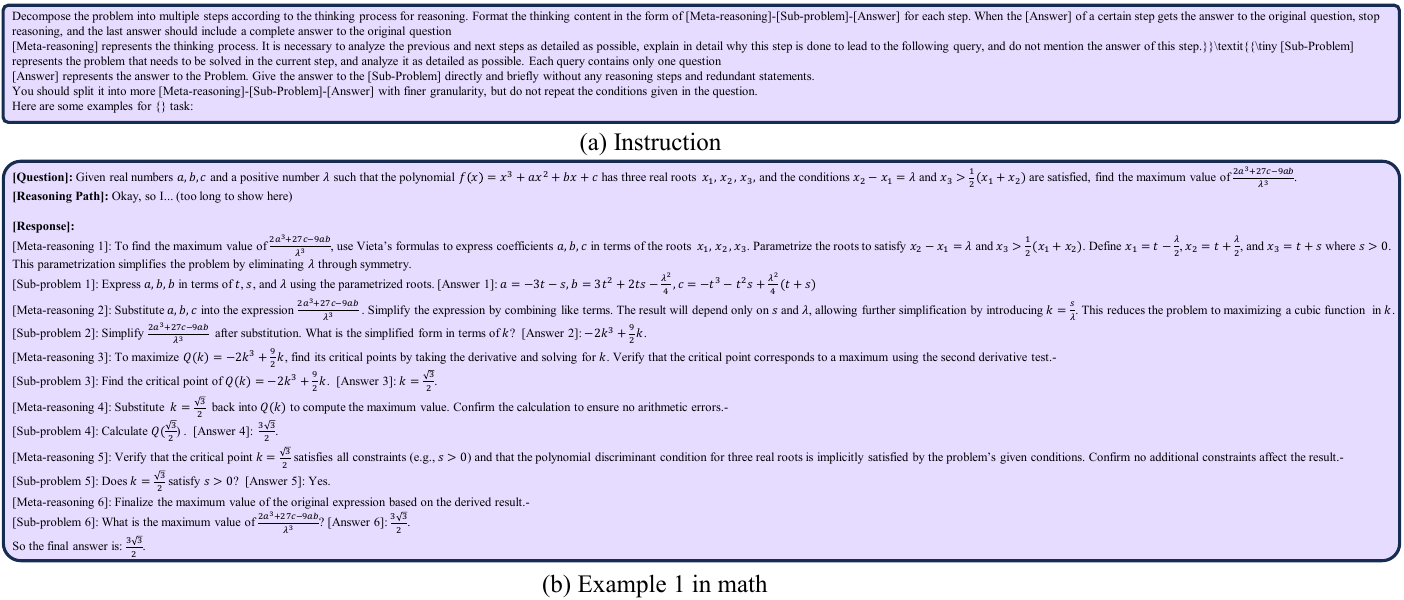}
        \caption{Full prompt in data construction of training generative structure reward model: Instruction and Example 1}
        \label{app_prompt}
\end{figure}

\begin{figure}[h]
    \centering
        \includegraphics[width=1.0\linewidth]{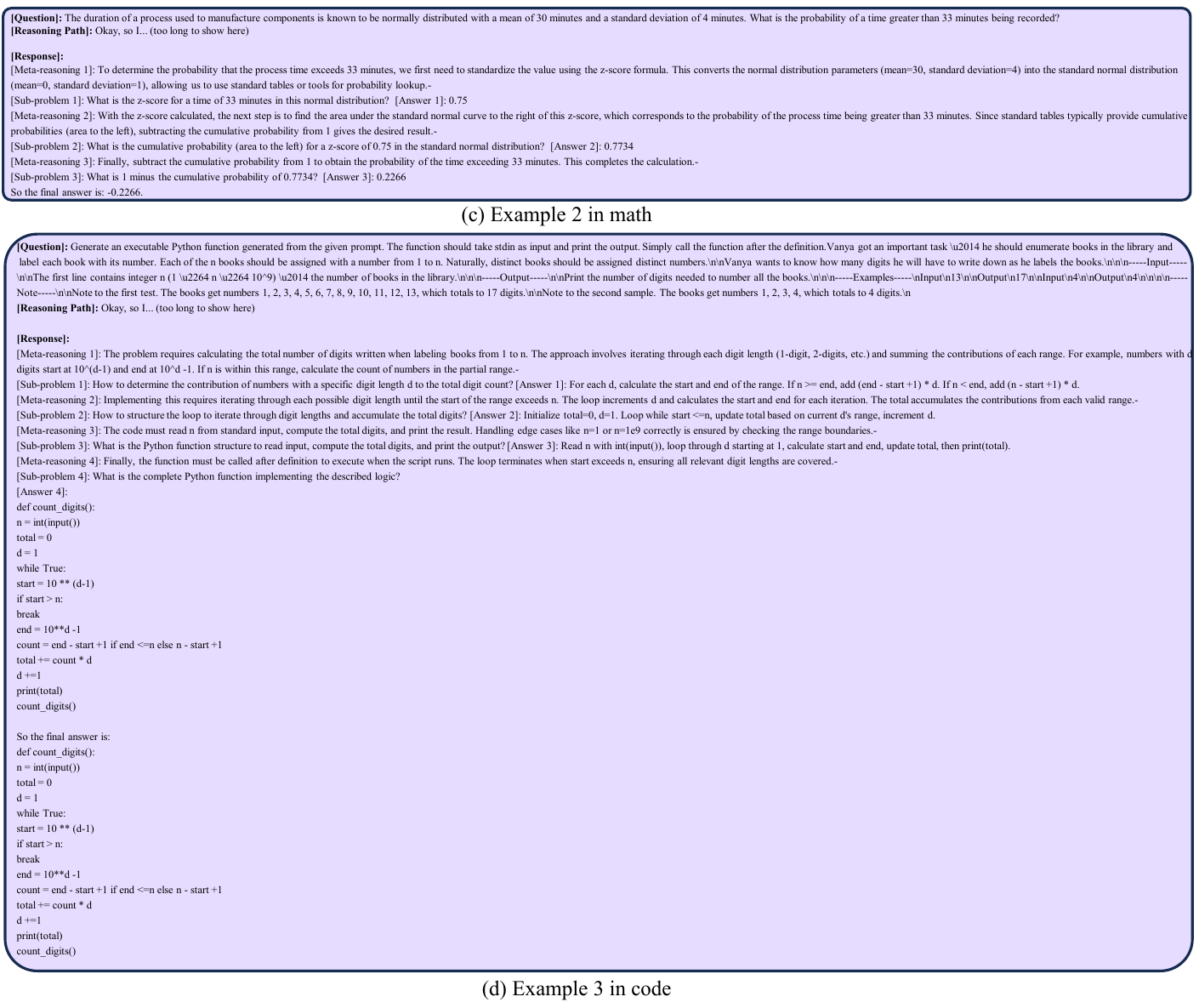}
        \caption{Full prompt in data construction of training generative structure reward model: Example 2 and 3}
        \label{app_prompt2}
\end{figure}

\begin{figure}[h]
    \centering
        \includegraphics[width=1.0\linewidth]{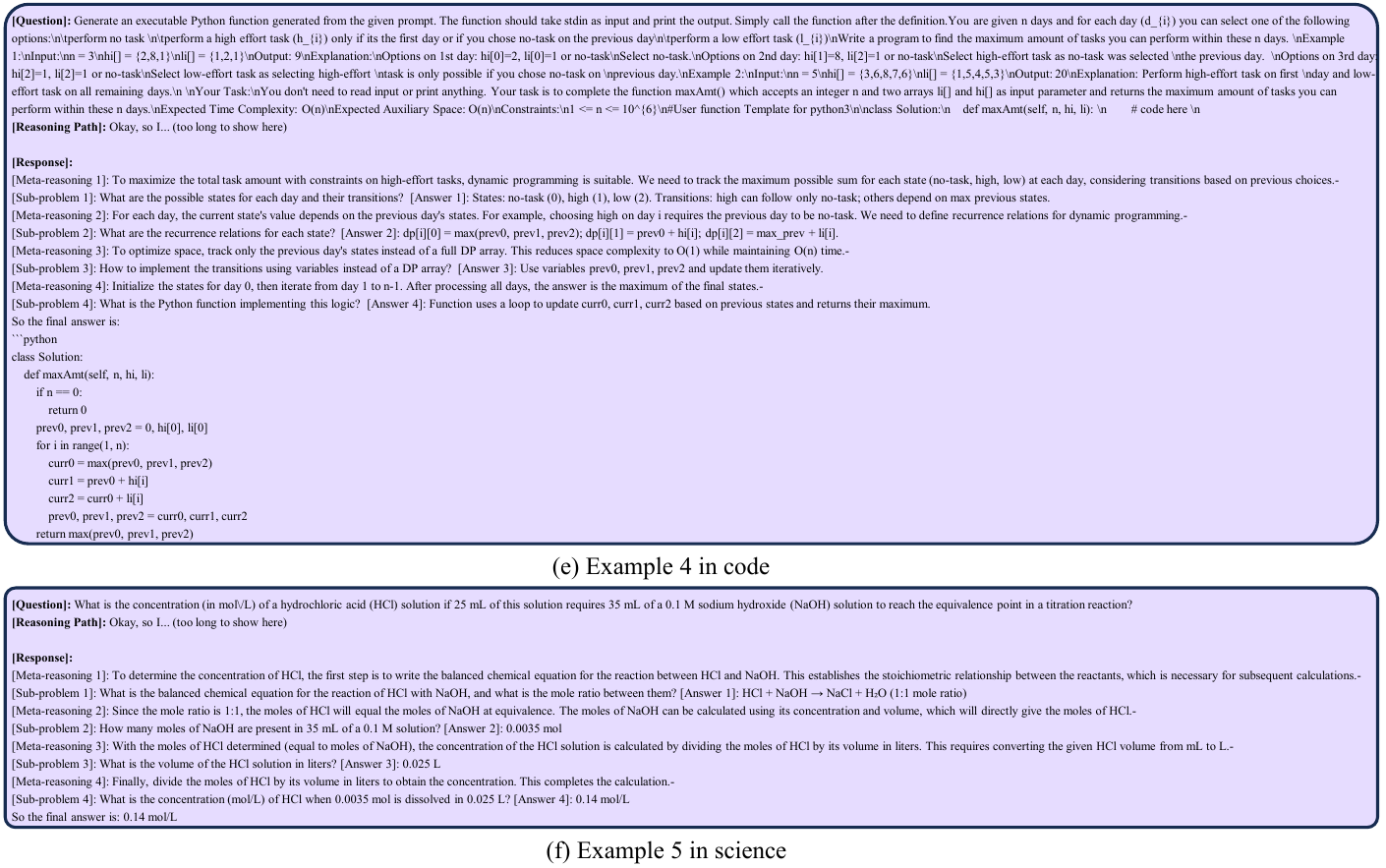}
        \caption{Full prompt in data construction of training generative structure reward model: Example 4 and 5}
        \label{app_prompt3}
\end{figure}

\begin{figure}[h]
    \centering
        \includegraphics[width=1.0\linewidth]{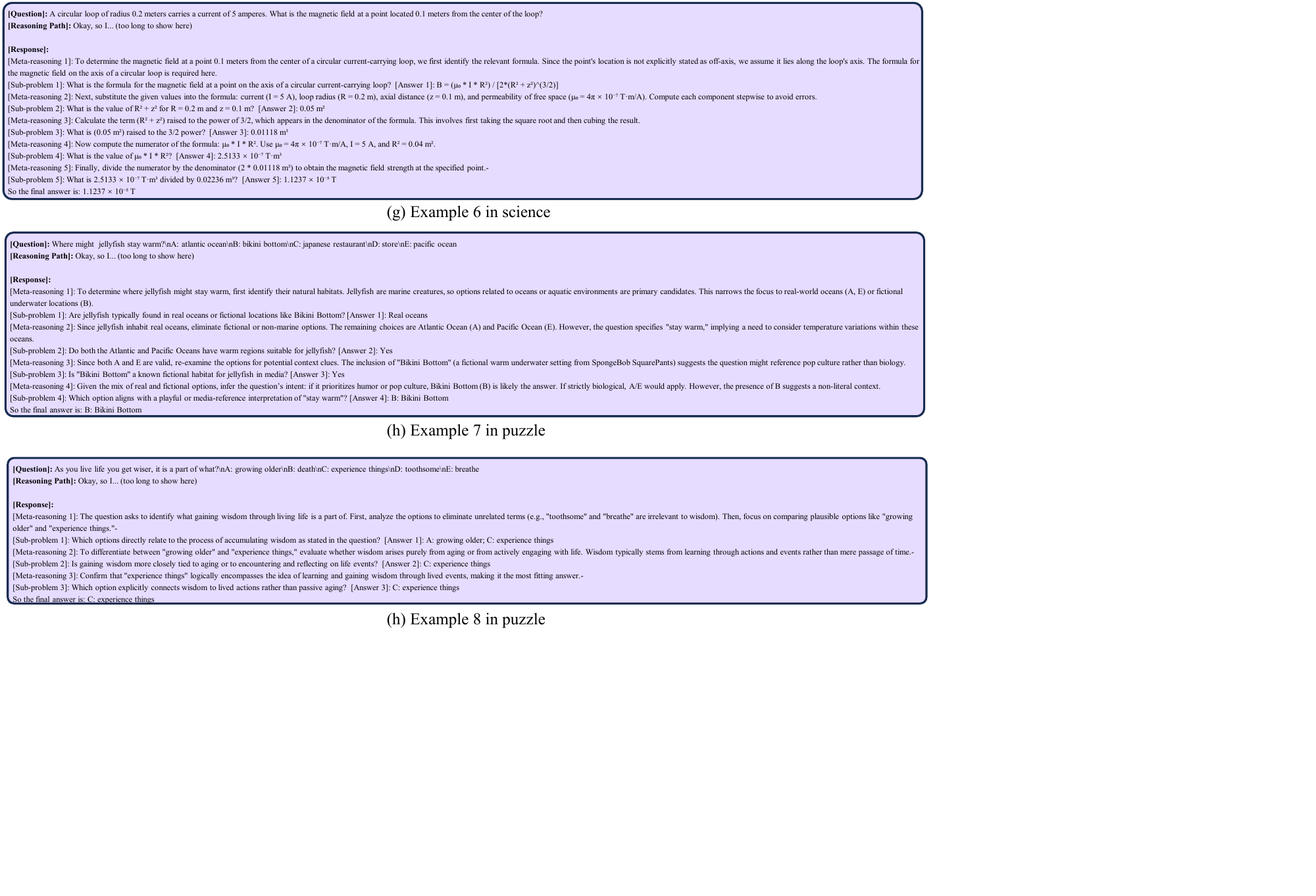}
        \caption{Full prompt in data construction of training generative structure reward model: Example 6, 7 and 8}
        \label{app_prompt4}
\end{figure}

Figure~\ref{app_prompt} to~\ref{app_prompt4} shows the full prompt in data construction of training generative structure reward model.

\end{document}